\documentclass[runningheads]{llncs}

\usepackage{eccv}

\usepackage{eccvabbrv}

\usepackage{graphicx}
\usepackage{booktabs}

\usepackage[accsupp]{axessibility}  

\usepackage{hyperref}

\usepackage{orcidlink}

\usepackage{multirow}
\usepackage{color}
\definecolor{gray}{RGB}{60,60,60}
\usepackage{makecell}

\begin{document}

\title{AnomalyFactory: Regard Anomaly Generation \\ 
as Unsupervised Anomaly Localization}

\titlerunning{AnomalyFactory}

\author{Ying Zhao\inst{1}\orcidlink{0000-0002-4113-8423} }

\authorrunning{Ying Zhao}

\institute{Ricoh Software Research Center (Beijing) Co., Ltd., China \\
\email{zy\_deepwhite\_zy@hotmail.com}}

\maketitle
\vspace{-10mm} 
\begin{figure}[htbp]
  \centering
  \includegraphics[height=3.3cm]{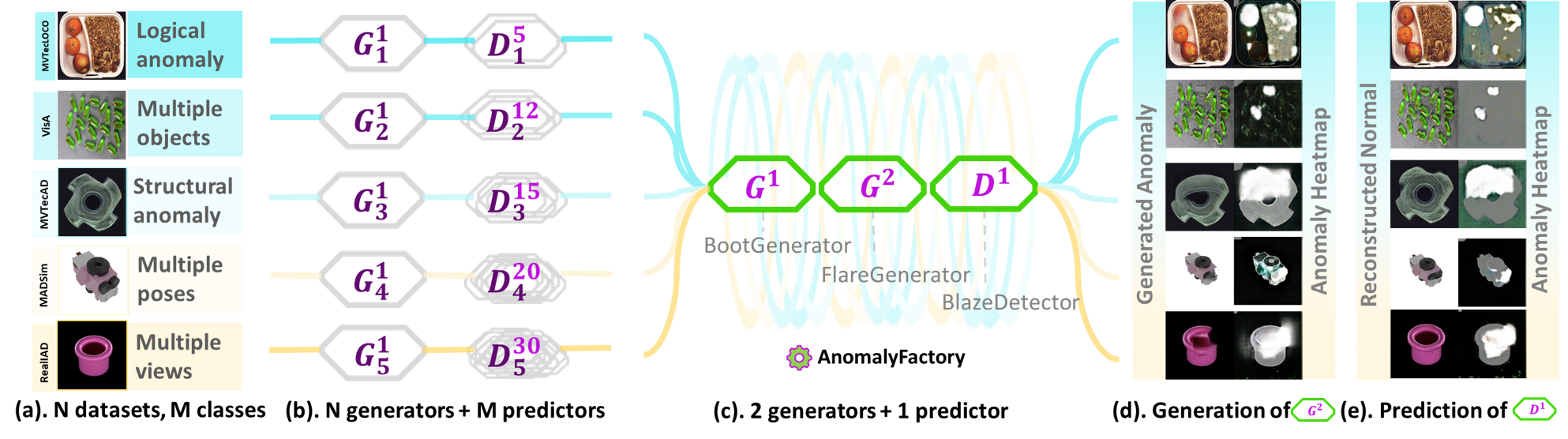}
  \caption{AnomalyFactory is a novel scalable framework that unifies unsupervised anomaly generation and localization. Unlike prior art, it learns only 2 generators and 1 predictor with same network architecture to solve tasks on M classes of N datasets. 
}
  \label{fig:fig0}
\end{figure}
\vspace{-10mm} 

\begin{abstract}
Recent advances in anomaly generation approaches alleviate the effect of data insufficiency on task of anomaly localization. 
While effective, most of them learn multiple large generative models on different datasets and cumbersome anomaly prediction models for different classes. 
To address the limitations, we propose a novel scalable framework, named AnomalyFactory, that unifies unsupervised anomaly generation and localization with same network architecture. It starts with a BootGenerator that combines structure of a target edge map and appearance of a reference color image with the guidance of a learned heatmap. Then, it proceeds with a FlareGenerator that receives supervision signals from the BootGenerator and reforms the heatmap to indicate anomaly locations in the generated image. Finally, it easily transforms the same network architecture to a BlazeDetector that localizes anomaly pixels with the learned heatmap by converting the anomaly images generated by the FlareGenerator to normal images. By manipulating the target edge maps and combining them with various reference images,  AnomalyFactory generates authentic and diversity samples cross domains. Comprehensive experiments carried on 5 datasets, including MVTecAD, VisA, MVTecLOCO, MADSim and RealIAD, demonstrate that our approach is superior to competitors in generation capability and scalability.

\end{abstract}

\section{Introduction}
\label{sec:intro}
Anomaly localization technology appeals to broad attention for its strong developing potential in industrial manufacturing. Despite of the increasing prosperousness, it remains constricted by the lack of real and diversity data. Most of recent methods partially mitigate the effect of data insufficiency by using extra large generative models. With the increased amounts of object categories from different datasets, their generation and localization models bring more and more heavy training cost and storage burden. Therefore, it is highly demanded a scalable framework that unifies anomaly generation and anomaly localization.

Fig.\ref{fig:fig0} illustrates an overview of proposed AnomalyFactory(Fig.\ref{fig:fig0}c) handling 82 object categories of 5 datasets(Fig.\ref{fig:fig0}a) comparing with prior art. Most existing methods \cite{realnet2024, anoDiffusion2024, DFMGAN2023} focus on learning a dedicated generative model for each dataset and M particular anomaly predictors for M categories. AnomalyDiffusion \cite{anoDiffusion2024} learns a 1.33GB-sized\footnote{https://github.com/sjtuplayer/anomalydiffusion} diffusion model and RealNet \cite{realnet2024} trains a 2.1GB-sized\footnote{https://github.com/cnulab/RealNet} diffusion model on MVTecAD\cite{mvtec} dataset for anomaly generation. DFMGAN \cite{DFMGAN2023} learns a 305.2MB-sized\footnote{https://github.com/Ldhlwh/DFMGAN} GAN model only for the hole defect of 1/15 of the MVTecAD categories(hazelnut). However, the real applications of industrial manufacturing always expect the models to have capability of handling large number of categories more effectively. Unlike the previous generation per dataset scheme shown in Fig.\ref{fig:fig0}b, our approach(Fig.\ref{fig:fig0}c) effectively learns all data from different domains together with only 2 generators and 1 predictor having 719MBx3 size.
The consecutive two generators, named BootGenerator and FlareGenerator, gradually improve the generation quality of anomaly images and heatmaps in an unsupervised way. This strategy makes training the anomaly predictor, named BlazeDetector, with the generated images possible.

More specifically, the BootGenerator combines structure of a target edge map and appearance of a reference color image with the guidance of a learned heatmap. The FlareGenerator receives the supervision signals from the BootGenerator on-the-fly and reforms the weight map to indicate anomaly locations in the generated image. The BlazeDetector localizes anomaly pixels with the learned heatmap by converting the anomaly images generated by the FlareGenerator to normal images.
With same network architecture, we easily transfer the function from anomaly generation to anomaly localization by switching the input normal reference image with the output anomaly image. Similar with FlareGenerator, our BlazeDetector is a unified model trained on all categories from different datasets. As shown in Fig.\ref{fig:fig0}e, our predictor can output a heatmap(right) accompany with the generated normal image(left) to indicate the pixel-level anomaly locations. 

\begin{figure}[tb]
  \centering
  \includegraphics[height=5.0cm]{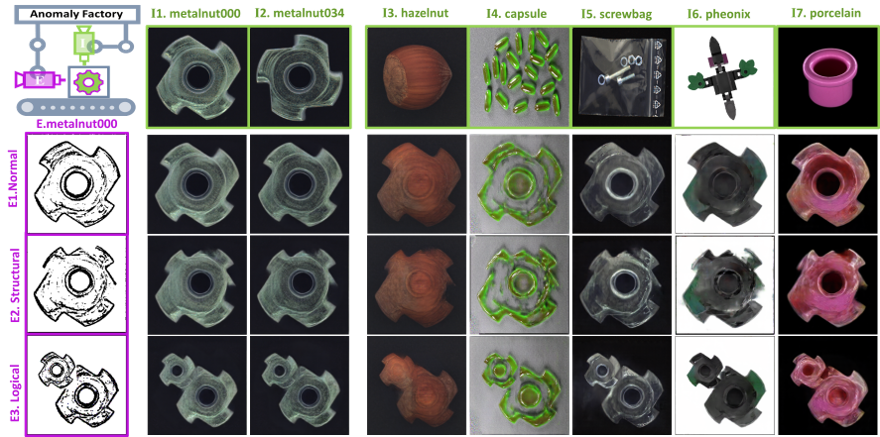}
  \caption{Illustrations of scalable anomaly generation of proposed FlareGenerator. The reference color image(I) provides texture content for the generation while the target edge map(E) controls the skeleton of generated object. By manipulating the target edge maps, our FlareGenerator generates normal(E1) samples and structural(E2) and logical(E3) anomalies. We further demonstrate its scalable ability by using reference color images from 5 different datasets, including MVTecAD\cite{mvtec}(I1-I3), VisA\cite{visa}(I4), MVTecLOCO\cite{mvtecloco}(I5), MADSim\cite{pad}(I6) and RealIAD\cite{realiad}(I7).}
  \label{fig:fig1}
\end{figure}

Besides the high-cost architecture, existing generative models focus only on either increasing diversity of structural anomalies \cite{realnet2024, anoDiffusion2024} or normal samples \cite{textguided2024} but ignore the logical anomaly \cite{mvtecloco} generation. On the contrary, the proposed FlareGenerator has strong scalability in generating various types of samples with anomaly heatmaps for training the anomaly predictor, as shown in Fig.\ref{fig:fig0}d and Fig.\ref{fig:fig1}. FlareGenerator generates scalable samples by combining features of the input target edge map(Fig.\ref{fig:fig1}E) and reference color image(Fig.\ref{fig:fig1}I).

The main contributions of this paper can be summarized as follows:

\begin{itemize} 
  \item We propose a novel scalable framework, named AnomalyFactory, that unifies unsupervised anomaly generation and localization. It learns only 2 generators and 1 predictor with same network architecture to solve tasks on M classess normal samples of N datasets.
  \item We design a GAN-based network architecture that combines structure of a target map and appearance of a reference color image with the guidance of a learned heatmap. It has strong scalability in generating various types of samples with anomaly heatmaps for training an unified anomaly predictor for multiple categories of different datasets.
  \item Extensive experiments carried on 82 object categories of 5 datasets demonstrate the superiority of AnomalyFactory over the existing anomaly generators for anomaly localization task. Our generator produces more realistic and diversity images. On MVTecAD \cite{mvtec} dataset, it achieves the highest 4.24 IS score comparing to SOTA and improves 2.37/\textcolor{gray}{0.11} higher IS/\textcolor{gray}{LPIPS} than the baseline with only a unified generator.
\end{itemize}

\section{Related Works}
\textbf{Anomaly Synthesis.}
To mitigate shortage of real anomaly samples, many unsupervised anomaly detection methods \cite{cutpaste, draem, jnld, omnial, slsg, visa} are closely paired with lost-cost anomaly synthesis.
CutPaste \cite{cutpaste} syntheses anomaly by cutting a local rectangular region from the normal image and paste it back at a random position. SPD \cite{visa} uses a smoothed version of CutPaste \cite{cutpaste} augmentation.
To increase the anomaly diversity, Draem \cite{draem} extracts anomaly texture from an extra DTD \cite{dtd} dataset and fuses it in the anomaly regions produced by using binarized Perlin noise.
To get a more concentrated anomaly region mask, SLSG \cite{slsg} extends Draem \cite{draem} by controlling the parameters of Perlin noise and the binarization threshold. 
Based on the just noticeable distortion \cite{jnd}, JNLD \cite{jnld} proposes a multi-scale noticeable anomalous generation method to simulate different levels of anomaly.
OmniAL \cite{omnial} further improves the anomaly synthesis of JNLD \cite{jnld} by using a panel-guided strategy to control the portion of synthetic anomaly. 
Increasingly, though, there are other supervised \cite{SDGAN2020, defectGAN2021, DFMGAN2023, anoDiffusion2024} and unsupervised \cite{realnet2024, grad2024, textguided2024} ways to generate more realistic and complex anomalies.

\begin{figure}[tb]
  \centering
  \includegraphics[height=3.5cm]{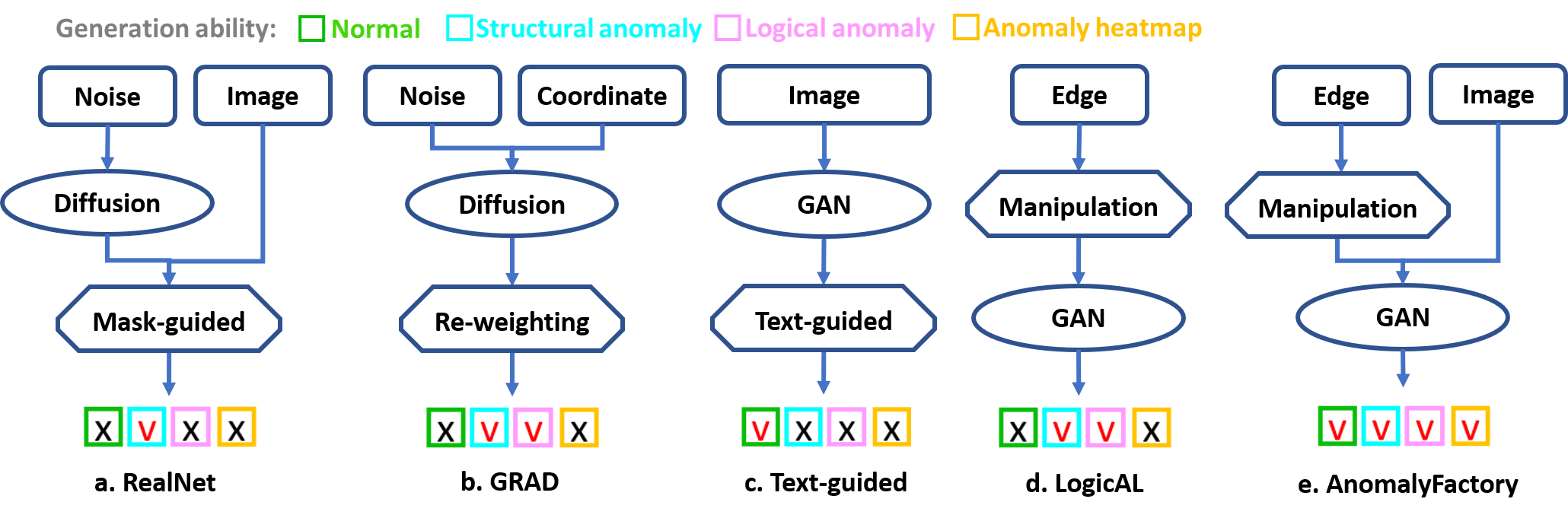}
  \caption{Comparison with unsupervised anomaly generation. We compare AnomalyFactory with prior arts \cite{realnet2024, grad2024, textguided2024, logical} from aspects of architecture and generation ability.}
  \label{fig:fig2}
\end{figure}

\noindent\textbf{Supervised Anomaly Generation.}
SDGAN \cite{SDGAN2020} converts defect-free to surface defects with the generator trained by cycle consistency loss on normal and anomaly images. DefectGAN \cite{defectGAN2021} generates anomaly samples by adding the learnt defect foregrounds on different normal backgrounds.
Both works require real defect images for training and neither of them provide pixel-level annotation for anomaly localization task.
DFMGAN \cite{DFMGAN2023} proposes a few-shot defect image generation method that generates structure anomalies based on a handful of defect samples. It attaches defect-aware residual blocks to the pre-trained StyleGAN2 \cite{stylegan2} backbone and manipulates the features within the learnt defect masks. AnoDiffusion \cite{anoDiffusion2024} proposes a diffusion-based few-shot anomaly generation model which utilizes the strong prior information of latent diffusion model learnt from large-scale dataset to enhance the generation authenticity.
These supervised anomaly synthesis methods, though effective, rely on real anomalous images and cannot generate unseen anomaly types.

\noindent\textbf{Unsupervised Anomaly Generation.}
RealNet \cite{realnet2024} proposes a diffusion process-based synthesis strategy that generates anomaly samples by blending the normal image with the diffusion generated anomalous texture. To mimic real anomalies distribution, it carefully selects the parameter that controls the strength of anomaly generation.
Although above mentioned methods are effective in structural anomaly synthesis, few of them could generate logical anomaly.
GRAD \cite{grad2024} proposes a diffusion model to generate both structural and logical anomaly patterns. It generates contrastive patterns by preserving the local structures while disregarding the global structures present in normal images. Moreover, it uses a self-supervised re-weighting mechanism to handle the challenge of long-tailed and unlabeled synthetic contrastive patterns. 
TextGuided \cite{textguided2024} utilizes text information about the target object, learned from extensive text library documents, to generate non-defective data images resembling the input image. 
LogicAL \cite{logical} proposes an edge manipulation based framework to produce photo-realistic both logical and structural anomalies. By modifying edges in semantic regions, it easily generates anomalies that break logical
constraints, such as missing components. By editing edges in arbitrary regions, it forges varies structural defects.
As shown in Fig.\ref{fig:fig2}, different with the aforementioned methods\cite{realnet2024, grad2024, textguided2024, logical}, we propose a novel approach that can generate both photo-realistic logical and structural anomalies and multiplicate normal samples cross different domains.

\noindent\textbf{Embedding-based Anomaly Detection.}
SDGAN \cite{SDGAN2020} trains the VGG \cite{vggloss} with its generated defect samples to recognize and classify anomaly.
DefectGAN \cite{defectGAN2021} inspects defect with the ResNet \cite{resnet} trained with its generated defect samples and the original dataset.
PatchCore \cite{patchcore} builds the memory bank with the sampled patch-level embeddings extracted by WideResNet \cite{wideresnet}. It detects abnormal samples based on the distance between the test sample's nearest neighbor feature in its memory bank and other features.
CutPaste \cite{cutpaste} fine-tunes the pretrained  EfficientNet(B4) \cite{efficientnet} on its synthetic anomaly dataset. With the fine-tuned feature extractor, it trains the Gaussian density estimator to compute anomaly score.
EfficientAD \cite{efficientad} uses the same pretrained features as PatchCore \cite{patchcore} from a WideResNet \cite{wideresnet}. It trains a patch description network on images from ImageNet \cite{imagenet} by minimizing the mean squared difference between its output and the features extracted from the pretrained network.

\noindent\textbf{Reconstruction-based Anomaly Detection.}
Draem \cite{draem} constructs a reconstructive and a discriminative subnetworks with the encoder-decoder architecture. With the subnetworks, it predicts the abnormal regions by calculating the difference between the reconstructed and original images.
To detect anomalies from different targets only using one model, OmniAL \cite{omnial} extends Draem \cite{draem} subnetworks with dilated channel and spatial attention blocks and DiffNeck module.
LogicAL \cite{logical} further explores the logical anomaly detection capability of OmniAL \cite{omnial} by learning edge information.
SLSG [29] uses a generative pre-trained network to learn the feature embedding of normal images. To model position information in images, it uses a self-supervised task to learn the reasoning of position relationships and use the graph convolutional network to capture across-neighborhood position relationships.
AnoDiffusion \cite{anoDiffusion2024} simply uses a UNet to detect anomaly and a resnet18 to classify anomaly type.
RealNet \cite{realnet2024} introduces anomaly-aware features selection and reconstruction residuals selection to improve anomaly detection performance.
GRAD \cite{grad2024} proposes a lightweight FCN-based patch-level detector to efficiently distinguish the normal patterns and generated anomaly patterns.

\begin{figure}[tb]
  \centering
  \includegraphics[height=6.0cm]{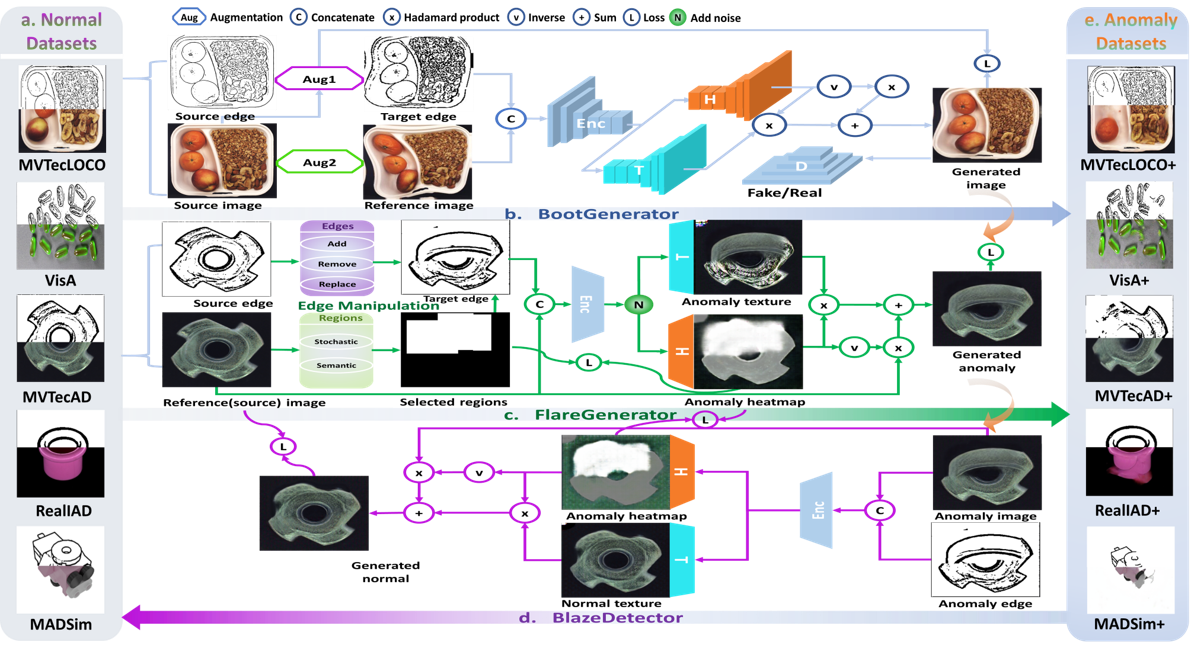}
  \caption{\textbf{Framework of AnomalyFactory.} It learns a unified generator that works for 5 datasets(a). The BootGenerator(b) triggers the generation ability of absorbing structure of a target edge map and appearance of a reference color image. The FlareGenerator(c) generates diversity anomaly images(e) accompany with more accurate anomaly heatmaps. By simply swapping normal images with the generated anomaly images for training, we learn the BlazeDetector(d) that directly outputs anomaly heatmaps.}
  \label{fig:fig3}
\end{figure}

\section{Method}
As illustrated in Fig.\ref{fig:fig3}, the proposed AnomalyFactory aims to efficiently generate and localize anomaly by learning from normal samples cross different datasets(Fig.\ref{fig:fig3}a). We design a unified network architecture that solves both tasks. To train the network only with normal samples, we introduce a progressive pipeline that consecutively learns a BootGenerator(Fig.\ref{fig:fig3}b), a FlareGenerator(Fig.\ref{fig:fig3}c) and a BlazeDetector(Fig.\ref{fig:fig3}d). The anomaly edge maps and corresponding anomaly images are shown in Fig.\ref{fig:fig3}e. The BootGenerator(Fig.\ref{fig:fig3}b) receives an target edge map and a reference color image having different augmentations and generates an image combining the structure and appearance of them. The FlareGenerator(Fig.\ref{fig:fig3}c) in next stage receives supervision signals from the BootGenerator and reforms the weight map to indicate anomaly locations in the generated image. The BlazeDetector(Fig.\ref{fig:fig3}d) in the final stage produces high quality anomaly heatmaps by learning anomaly images generated by the FlareGenerator.

\subsection{Network Architecture}
Our aim is to build a unified network architecture that works for anomaly generation and anomaly localization. Essentially, these two tasks both can be defined as following formulation.
\begin{equation}
  I_{out} = I_{in}\cdot(1-H)+T \cdot H
  \label{eq:important}
\end{equation}
Where the output image $I_{out}$ is a fusion of the input image $I_{in}$ and the generated texture image $T$ with the guidance of a weight map $H$. In terms of anomaly generation task, the input $I_{in}$ is a normal image and the output $I_{out}$ is an anomaly image. Thus, the generated texture image $T$ is the source of final anomaly and the fusion weight map $H$ indicates the pixel-level anomaly locations. By swapping the content of $I_{in}$ and $I_{out}$,we obtain the formulaic representation of anomaly localization task. In this case, the texture image $T$ provides the normal ingredient and the weight map $H$ is the prediction result for anomaly localization task. According to the ablation study of LogicAL \cite{logical}, the generation network taking only edge map as input has mode collapse problem and limited diversity. To make our network be scalable, we further improve the network design to take a target edge map with a reference color image as input. By manipulating the target edge map and combining it with different reference color image, we can largely increase the generation diversity.

Motivated by this observation, we design a network architecture  based on the pix2pixHD \cite{pix2pixhd} that is a conditional generative adversarial network (cGAN) models the conditional distribution of real images given the input edge maps via the minimax game. The pix2pixHD \cite{pix2pixhd} consists of a coarse-to-fine generator translating edge maps to color images and a multi-scale discriminator distinguishing real images from the generated ones. We modify the network input to receive a concatenation of the target edge map and the reference color image. Inspired by GANimation \cite{ganimation}, we tailor the generator into separate heatmap decoder and texture decoder followed by a non-parameters fusion module, as shown in Fig.\ref{fig:fig3}.

When training the model, we do not have paired real anomaly images and ground-truth annotation for anomaly texture images and heatmaps. Therefore, we design a progressive training pipeline that focuses on one of the learning targets in each stage. In the first stage, we train a BootGenerator that concentrates on stably generating images combining structure of the target edge map and appearance of the reference color image. Given the the generated images to be ground-truth anomaly in the second stage, we train a FlareGenerator that focuses on improving the quality of anomaly heatmap generation. Lastly, with the anomaly image and heatmap generated by FlareGenerator, we train an anomaly predictor BlazeDetector .

\subsection{BootGenerator}
The key ingredient of training BootGenerator is to make proper difference between the input target edge map and reference color image through augmentations. This strategy prevents the generator learn a shortcut to generate all zeros or all ones fusion weight map. Thanks to this strategy, the generated image absorbs features from both inputs.

\begin{figure}[tb]
  \centering
  \includegraphics[height=4.cm]{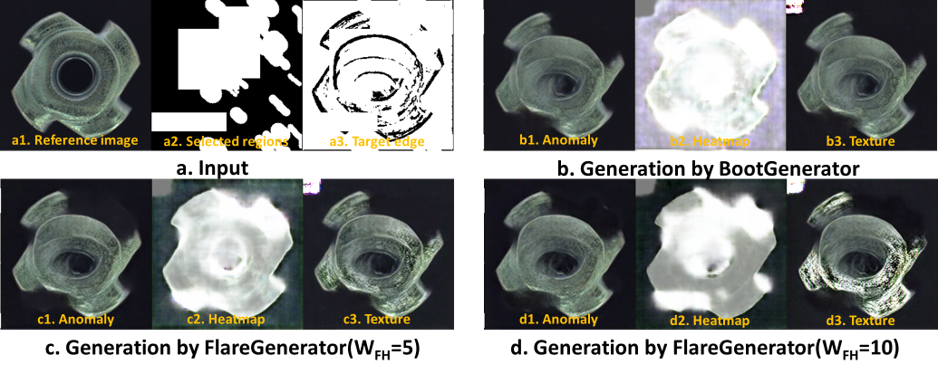}
  \caption{Compare BootGenerator with FlareGenerator.}
  \label{fig:fig4}
\end{figure}

\textbf{Augmentations.} As shown in Fig.\ref{fig:fig3}b, the source inputs are a pair of edge map and color image. We apply different augmentations on them to build a target edge map and a reference color image. To build the target generated image, we also apply the same augmentation of edge map on the color image. These augmentations mainly consist of three types, including local thin-plate-spline (TPS) \cite{tps} warps, resize-translation-padding and top-bottom/left-right flip. The local TPS randomly shifts 3x3 control points from a local region in the horizontal and vertical directions. Compare to selecting control points from whole image, the local TPS brings smaller spatial range of warps. The resize-translation-padding augmentation is mainly design to counter the drastically edge manipulation on texture categories, such as editing edges of the most parts of splicingConnectors \cite{mvtecloco} category. The flip augmentation brings the benefits of direction-agnostic authentic generation. By using these three augmentations, we get a generator that is robust to the drastically edge manipulation, as the example shown in Fig.\ref{fig:fig4}a3.

\textbf{Losses.} Following DeepSIM \cite{deepsim}, we use the VGG perceptual loss \cite{vggloss} $L_{vgg}$ to measure the fidelity of generation $G(E_t,I_r)$ and the conditional GAN loss $L_D$ to measure the differentiate between the generated and true images. The total loss $L_{BG}$ of BootGenerator is defined as follow.
\begin{equation}
  L_G(E_t,I_r,I_t;G) = L_{vgg}(G(E_t,I_r), I_t)
  \label{eq:Lg}
\end{equation}
\vspace{-8mm}
\begin{equation}
  L_D(E_t,I_r,I_t;D,G) = log(D(E_t,I_r,I_t))+log(1-D(E_t,I_r,G(E_t,I_r)))
  \label{eq:Lg}
\end{equation}
\vspace{-5mm}
\begin{equation}
  L_{BG} = L_G(E_t,I_r,I_t;G) + L_D(E_t,I_r,I_t;D,G)
  \label{eq:Lbg}
\end{equation}
Where, $E_t$ is the input target edge map, $I_r$ is the input reference color image, $I_t$ is the target image of generation, $G$ is the generator and $D$ is the discriminator of BootGenerator.

As the example shown in Fig.\ref{fig:fig4}b2, the heatmap generated by BootGenerator is very noise and barely can accurately indicate the anomaly pixels. On the contrary, the heatmap(Fig.\ref{fig:fig4}d2) generated by the FlareGenerator is better aligned with the true anomaly pixels.

\subsection{FlareGenerator}
The main target of FlareGenerator is to generate anomaly images accompany with anomaly heatmaps that accurately indicate pixel-level anomaly locations.

\textbf{Edge Manipulation.} As shown in Fig.\ref{fig:fig3}c, how to forge anomaly edges is the fountainhead of anomaly generation. The edge manipulation module consists of region selection and edge editing. Following LogicAL \cite{logical}, the region selection consists of semantic and stochastic region selection strategies. We use the pre-trained segmentation model SAM \cite{sam} to get rough semantic regions. Due to SAM \cite{sam} is over-segmented, we further refine the segmentations by removing background, grouping small regions and merging overlapped regions and build a map of candidate regions. For stochastic region selection, we randomly combine multiple regions with different aspect ratios or shapes. Given the selected regions, we modify the normal edges by removing, replacing or merging strategies.
Unlike LogicAL \cite{logical}, we don't need to carefully restrict the number of selected regions and the edge editing strategies. The candidate anomaly edges can be derived from any categories. As shown in Fig.\ref{fig:fig4}, FlareGenerator can generate high quality anomaly image even if the target edge map is drastically edited. Thanks to the edge manipulation, FlareGenerator generates diversity structural and logical anomalies, as illustrated in Fig.\ref{fig:fig3}e. By adding random noise to the encoded features, the generated images using same normal edge map and color image can also be different from each other.

\textbf{Losses.}
The ground truth images $I_t$ are generated by BootGenerator. Besides the VGG perceptual loss $L_{vgg}$ to measure the fidelity of generation $G(E_t,I_r)$ and the conditional GAN loss $L_D$, we use L2 loss to measure the quality of anomaly heatmap $H$ with the supervision of select region mask $M$. The total loss $L_{FG}$ of FlareGenerator is defined as follow.
\begin{equation}
  L_{FH} = \dfrac{1}{n}\sum_{i=1}^{n}(H_i-M_i)^2
  \label{eq:Lh}
\end{equation}
\vspace{-5mm}
\begin{equation}
  L_{FG} = W_{FH}*L_{FH} + L_G(E_t,I_r,I_t;G) + L_D(E_t,I_r,I_t;D,G)
  \label{eq:Lfg}
\end{equation}

Where, $W_{FH}$($W_{FH}=10$) is the weight of anomaly heatmap loss $L_{FH}$.

\subsection{BlazeDetector}
Limited by anomaly generation quality, the prior art \cite{logical, omnial} have to calculate the ground truth of anomaly heatmap with SSIM \cite{ssim} that provides a measure of the similarity by comparing two images based on luminance similarity, contrast similarity and structural similarity information. On the contrary, our BlazeDetector learns the anomaly heatmap directly generated by FlareGenerator. As shown in Fig.\ref{fig:fig3}d, BlazeDetector receives a pair of anomaly edge map and color image and outputs a reconstructed normal image accompany with a anomaly heatmap.  
Similar to FlareGenerator, we use L2 loss to measure the quality of anomaly heatmap $BH$ with the supervision of heatmap $FH$ generated by FlareGenerator.
The total loss $L_{BD}$ of BlazeDetector is defined as follow.
\begin{equation}
  L_{BD} = W_{BH}*L_{BH} + L_G(E_t,I_r,I_t;G) + L_D(E_t,I_r,I_t;D,G)
  \label{eq:Lbd}
\end{equation}
Where, $W_{BH}$($W_{BH}=10$) is the weight of anomaly heatmap loss $L_{BH}$.

\section{Experiments}
To demonstrate the effectiveness of proposed AnomalyFactory, we conduct extensive experiments on the challenging MVTecLOCO \cite{mvtecloco}, MVTecAD \cite{mvtec}, VisA \cite{visa}, MADSim \cite{pad} and RealIAD \cite{realiad} datasets. We evaluate the overall performance of anomaly generation and localization comparing with existing advanced methods. Our experiments also include ablation studies on the effectiveness of each component.
\subsection{Datasets}
As definition in MVTecLOCO \cite{mvtecloco}, industrial anomalies can be organized into structural and logical types. The structure anomalies contain various flaws, including surface defects such as scratches, dents, color spots or crack. The logical anomalies contain flawless components that violate underlying
logical constraints, such as misplacement or missing parts. Considering these two types of anomalies in various condition, our experiments are carried on 82 object categories from following five datasets. \textbf{MVTecLOCO} \cite{mvtecloco} dataset contains 5 categories of approximately 3,644 images covering both logical and structural anomalies from industrial inspection scenarios. \textbf{MVTecAD} \cite{mvtec} dataset contains 1,258 test images but pays more attention on structural anomalies than MVTecLOCO \cite{mvtecloco}. 
\textbf{VisA} \cite{visa} dataset consists of 9,621 normal and 1,200 anomalous color images covering 12 objects in 3 domains, including complex structure, multiple instances and single instances. 
\textbf{MADSim} \cite{pad} dataset contains 5,231 normal and 4,902 anomaly color images of 20 types of 3D LEGO animal models from different viewpoints covering a wide range of poses. 
\textbf{RealIAD} \cite{realiad} dataset contains 150K high-resolution images of 30 different objects captured in five views.

\subsection{Implementation}
To balance the amounts of different categories, we randomly sample 200x12 images from VisA\cite{visa} and 400x30 images from RealIAD\cite{realiad}. We totally use 18,007 normal images from 82 categories of 5 datasets for training. We extract edge maps and semantic region masks with pre-trained PiDiNet \cite{pidinet} and SAM \cite{sam} respectively in advance for training. Our network architecture and training schedule are based on Pix2PixHD \cite{pix2pixhd}. It mainly evolves four scales features encoding and decoding with basic convolution blocks and ResnetBlocks. The decoders for generating anomaly texture and heatmap have same structure of original decoder of Pix2PixHD. With the pairs of target edge maps and reference color images, we firstly train the BootGenerator 30 epochs with a batch size of 32 images having size of 256x256. With anomaly images generated by the BootGenerator on-the-fly, we train the FlareGenerator 5 epochs. Similarly, we train the BlazeDetector with pairs of normal images and anomaly images generated by FlareGenerator for 5 epochs. The local TPS augmentation is used only in training BootGenerator. The Adam optimizer has an initial learning rate of 2e-4 and decreases the learning rate with linear schedule.

\subsection{Metrics}
For generation evaluation, Kernel Inception Distance(KID), Frechet Inception Distance(FID), Inception Score(IS) and Learned Perceptual Image Patch Similarity(LPIPS) are four popular metrics. 
KID and FID measure the distance between the distribution of generated and real images. However, due to the limited amount of real anomaly data, KID and FID tend to yield better scores for the over-fitted models. IS measures the realistic and diversity of generated images but is independent of the given real anomaly data. A higher IS indicates better realistic and greater diversity.
LPIPS computes the similarity between the features of two image patches extracted from a pre-trained network. The higher LPIPS the greater variety generated images are. Following AnomalyDiffusion \cite{anoDiffusion2024}, we utilize IS and cluster-based LPIPS to evaluate the realistic and diversity of our generation.

\subsection{Performance}
Table \ref{table:1} shows the comparison in logical and structural anomalies generation using different region selection strategies between the baseline LogicAL\cite{logical} and our FlareGenerator. For both logical and structural anomalies generation, our method generates more realistic and diversity samples with higher IS and LPIPS scores. As shown in Fig.\ref{fig:fig7}, the semantic region selection(top row) produces complete object regions and the stochastic region selection outputs random regions(bottom row). Overall, the stochastic region selection brings more diversity than semantic region selection on MVTecLOCO \cite{mvtecloco} dataset. Considering the complexity of 5 datasets, we chose to mix these two region selection strategies for all the experiments.

\begin{table*}[t]
\begin{center}

\setlength{\abovecaptionskip}{0.cm}
\caption{Generation quality comparison of FlareGenerator with baseline method LogicAL \cite{logical} on logical and structural anomalies using different region selection strategies.} \label{table:1}
\resizebox{1.0\textwidth}{14mm}{
\begin{tabular}{c|ccc|ccc||ccc|ccc}

  \hline
  \multirow{4}{*}{\makecell[c]{\textbf{MVTec}\\\textbf{LOCO}}}  
  &\multicolumn{6}{c||}{\textbf{logical anomaly} Generation}
  &\multicolumn{6}{c}{\textbf{structural anomaly} Generation}

  \\
  \cline{2-13}
  & \multicolumn{3}{c|}{\textbf{LogicAL}}  
  & \multicolumn{3}{c||}{\textbf{FlareGenerator}}    
  & \multicolumn{3}{c|}{\textbf{LogicAL}} 
  & \multicolumn{3}{c}{\textbf{FlareGenerator}}

  \\
  \cline{2-13}
  & \multicolumn{1}{c}{\textbf{Semantic}}  
  & \multicolumn{1}{c}{\textbf{Stochastic}}    
  & \multicolumn{1}{c|}{\textbf{Mix}} 
  
  & \multicolumn{1}{c}{\textbf{Semantic}}  
  & \multicolumn{1}{c}{\textbf{Stochastic}}    
  & \multicolumn{1}{c||}{\textbf{Mix}} 
  
  & \multicolumn{1}{c}{\textbf{Semantic}}  
  & \multicolumn{1}{c}{\textbf{Stochastic}}    
  & \multicolumn{1}{c|}{\textbf{Mix}} 
  
  & \multicolumn{1}{c}{\textbf{Semantic}}  
  & \multicolumn{1}{c}{\textbf{Stochastic}}    
  & \multicolumn{1}{c}{\textbf{Mix}} 
  \\
  \cline{2-13} & \multicolumn{12}{c}{IS$\uparrow$ (Inception Score) / \textcolor{gray}{LPIPS\textbf{x10}$\uparrow$ (Learned Perceptual Image Patch Similarity)}}\\
    
    \hline
  breakfastB 
	&1.37/\textcolor{gray}{0.92}   
	&1.55/\textcolor{gray}{1.12}   	
	&1.46/\textcolor{gray}{0.99}

	&\textbf{4.51}/\textcolor{gray}{2.01}   
	&4.31/\textbf{\textcolor{gray}{2.31}} 
	&4.48/\textcolor{gray}{2.21}

	&1.37/\textcolor{gray}{0.81} 
	&1.54/\textcolor{gray}{1.16}	
	&1.44/\textcolor{gray}{0.90}

	&4.49/\textcolor{gray}{2.43} 
	&4.43/\textcolor{gray}{2.50}   	
	&\textbf{4.54/\textcolor{gray}{2.52}}

  		\\
  		
  juiceBottle 
	&1.38/\textcolor{gray}{1.37}    
	&1.55/\textcolor{gray}{1.79}   	
	&1.47/\textcolor{gray}{1.68}

	&3.83/\textcolor{gray}{3.19}   
	&\textbf{3.97}/\textcolor{gray}{3.30} 
	&3.94/\textbf{\textcolor{gray}{3.39}}

	&1.40/\textcolor{gray}{1.46} 
	&1.54/\textcolor{gray}{1.92}	
	&1.47/\textcolor{gray}{1.75}

	&3.80/\textcolor{gray}{3.00} 
	&3.96/\textcolor{gray}{3.33}   	
	&\textbf{4.00/\textcolor{gray}{3.41}} 

  		\\
  		
  pushpins 
	&1.23/\textcolor{gray}{3.25}   
	&1.32/\textbf{\textcolor{gray}{3.38}}   	
	&1.27/\textcolor{gray}{3.31}

	&3.77/\textcolor{gray}{3.18}   
	&\textbf{3.91}/\textcolor{gray}{3.35} 
	&3.84/\textcolor{gray}{3.20}

	&1.23/\textcolor{gray}{2.69}	
	&1.32/\textcolor{gray}{3.10} 
	&1.28/\textcolor{gray}{2.98} 
	
	&3.78/\textcolor{gray}{3.20} 
	&\textbf{3.95/\textcolor{gray}{3.38}}   	
	&3.90/\textcolor{gray}{3.28} 
  		\\
  		
  screwBag 
	&1.31/\textcolor{gray}{2.03}   
	&1.49/\textcolor{gray}{2.58}   	
	&1.47/\textcolor{gray}{2.31}

	&4.75/\textcolor{gray}{3.38}   
	&\textbf{5.34/\textcolor{gray}{3.65}}
	&5.06/\textcolor{gray}{3.45}

	&1.32/\textcolor{gray}{2.03} 
		&1.51/\textcolor{gray}{2.57}  
	&1.43/\textcolor{gray}{2.44}

	&4.91/\textcolor{gray}{3.39} 
	&5.03/\textbf{\textcolor{gray}{4.11}}   	
	&\textbf{5.04}/\textcolor{gray}{3.55} 
  		\\
  		
  splicingC
	&1.82/\textcolor{gray}{2.24}   
	&2.19/\textbf{\textcolor{gray}{2.30}}   	
	&1.96/\textcolor{gray}{2.28}

	&3.70/\textcolor{gray}{1.90}   
	&\textbf{4.02}/\textcolor{gray}{2.12}
	&2.81/\textcolor{gray}{1.98}

	&1.83/\textcolor{gray}{2.25} 
		&2.11/\textcolor{gray}{2.28} 
	&1.93/\textcolor{gray}{2.35}

	&3.64/\textcolor{gray}{2.30} 
	&\textbf{4.00/\textcolor{gray}{2.42}}   	
	&3.91/\textcolor{gray}{2.31} 
  	\\
  	
  \hline
  Mean 
	&1.42/\textcolor{gray}{1.96}    
	&1.62/\textcolor{gray}{2.23}   	
	&1.52/\textcolor{gray}{2.11}

	&4.11/\textcolor{gray}{2.73}   
	&\textbf{4.31/\textcolor{gray}{2.94}}
	&4.23/\textcolor{gray}{2.85}

	&1.43/\textcolor{gray}{1.85} 
		&1.60/\textcolor{gray}{2.20}
	&1.51/\textcolor{gray}{2.08}

	&4.12/\textcolor{gray}{2.87} 
	&4.27/\textbf{\textcolor{gray}{3.15}}   	
	&\textbf{4.28}/\textcolor{gray}{3.01} 

  	\\
  \hline
\end{tabular}
	}
\vspace{-0.5cm}
\end{center}
\end{table*}  

\begin{table*}[htbp]
\begin{center}
\setlength{\abovecaptionskip}{0.cm}
\caption{Generation quality comparison of FlareGenerator(FlareG) with baseline method LogicAL \cite{logical} and BootGenerator(BootG) on 5 datasets.} \label{table:2}
\resizebox{1.0\textwidth}{14mm}{
\begin{tabular}{c|cc|cc||ccccc|c}

  \hline
  \multirow{3}{*}{\textbf{Dataset}} 
  &\multicolumn{2}{c|}{\textbf{Normal} Generation}
  &\multicolumn{2}{c||}{\textbf{Anomaly} Generation}
  &\multicolumn{5}{c|}{Reference color images for \textbf{FlareG} are from}
  &\multicolumn{1}{c}{Mean}
  
  \\
  \cline{2-11}
  & \multicolumn{1}{c}{\textbf{LogicAL}}  
  & \multicolumn{1}{c||}{\textbf{FlareG}}    
  & \multicolumn{1}{c}{\textbf{LogicAL}} 
  & \multicolumn{1}{c||}{BootG}

  & \multicolumn{1}{c}{+mvtecad}  
  & \multicolumn{1}{c}{+mvtecloco}
  & \multicolumn{1}{c}{+visa}  
  & \multicolumn{1}{c}{+madsim}
  & \multicolumn{1}{c|}{+realiad}
  & \multicolumn{1}{c}{\textbf{FlareG}}
  \\
    \cline{2-11} & \multicolumn{9}{c}{IS$\uparrow$ (Inception Score) / \textcolor{gray}{LPIPS\textbf{x10}$\uparrow$ (Learned Perceptual Image Patch Similarity)}}\\
    
    \hline
  MVTecAD 
	&1.32/\textcolor{gray}{0.46} 
  	&\textbf{1.36}/\textbf{\textcolor{gray}{0.50}} 
  	 
  	&1.87/\textcolor{gray}{1.37} 
  	&\underline{5.50/\textcolor{gray}{1.94}}

  	&\underline{5.50/\textbf{\textcolor{gray}{1.96}}}  
  	&3.92/\textcolor{gray}{2.61}
  	&4.37/\textcolor{gray}{3.17}
  	&3.02/\textcolor{gray}{1.37}
  	&4.40/\textcolor{gray}{3.21}
  	
  	&\textbf{4.24}/\textbf{\textcolor{gray}{2.47}} 
  		 
  		\\
  		
  MVTecLOCO 
		&1.35/\textcolor{gray}{1.36}   		  		 
  		&\textbf{1.49}/\textbf{\textcolor{gray}{1.62}}
  		
  		&1.52/\textcolor{gray}{2.12} 
  		&\underline{2.90/\textcolor{gray}{2.60}}

  		&5.83/\textcolor{gray}{3.07}  
  		&\textbf{\underline{4.19/\textcolor{gray}{2.88}}}
  		&4.67/\textcolor{gray}{4.43}
  		&3.23/\textcolor{gray}{1.20}
  		&4.33/\textcolor{gray}{4.02}
  		 
  		&\textbf{4.45}/\textbf{\textcolor{gray}{3.12}}
  		\\
  		
  VisA 
		&1.25/\textcolor{gray}{1.03}   		 		 
  		&\textbf{1.44}/\textbf{\textcolor{gray}{1.74}}
  		 
  		&1.64/\textcolor{gray}{2.02}
  		&\underline{4.18/\textcolor{gray}{3.33}}

  		&-/\textcolor{gray}{-} 
  		&-/\textcolor{gray}{-}
  		&\textbf{\underline{4.50/\textcolor{gray}{3.59}}}
  		&-/\textcolor{gray}{-}  		
  		&\textbf{-}/\textbf{\textcolor{gray}{-}}
  		
  		&\textbf{4.50}/\textbf{\textcolor{gray}{3.59}}
  		\\
  		
  MADSim 
		&2.43/\textcolor{gray}{0.64}  		  		 
  		&\textbf{3.11}/\textbf{\textcolor{gray}{0.88}}
  		
  		&2.97/\textcolor{gray}{0.79}
  		&\underline{3.24/\textcolor{gray}{1.16}}

  		&-/\textcolor{gray}{-}  
  		&-/\textcolor{gray}{-}
  		&-/\textcolor{gray}{-}
  		&\textbf{\underline{3.27/\textcolor{gray}{1.12}}}
  		&-/\textcolor{gray}{-}
  		
  		&\textbf{3.27}/\textbf{\textcolor{gray}{1.12}} 
  		\\
  		
  RealIAD 
		&-/\textcolor{gray}{-}  
		&\textbf{1.98}/\textbf{\textcolor{gray}{1.45}} 
		 		
  	   &-/\textcolor{gray}{-}
  	   &\underline{5.59/\textcolor{gray}{3.84}}
  	     	    
  	   &-/\textcolor{gray}{-} 
  	   &-/\textcolor{gray}{-} 
  	   &-/\textcolor{gray}{-}
  	   &-/\textcolor{gray}{-}
  	   &\textbf{\underline{5.63/\textcolor{gray}{4.03}}}

		&\textbf{5.63}/\textbf{\textcolor{gray}{4.03}}
  	\\
  	
  \hline
  Mean 
	&1.59/\textcolor{gray}{0.87} 
  	&\textbf{1.88}/\textbf{\textcolor{gray}{1.19}} 
  	 
  	&2.00/\textcolor{gray}{1.58} 
  	&\underline{3.96/\textcolor{gray}{2.26}}
  	
  	& \multicolumn{5}{c|}{\textbf{\underline{4.62/\textcolor{gray}{2.72}}}}

  	&\textbf{4.42}/\textbf{\textcolor{gray}{2.58}} 
  	\\
  \hline
\end{tabular}
	}
\vspace{-0.5cm}
\end{center}
\end{table*}

As shown in Table \ref{table:2}, for the evaluation on each dataset, we conduct extensive anomaly generation experiments with reference color images and source anomaly edges from different datasets. For fair comparison, we build fixed lists of 1,000 randomly sampled normal images for each category of each training dataset. We also construct lists of 100 sampled normal and anomaly images from the testing dataset. To calculate LPIPS, we partition the generated 1,000 images into 100 groups by finding the lowest LPIPS. We compute the mean pairwise LPIPS within each group. The average LPIPS of all groups will be the final score. Our baseline method is LogicAL \cite{logical} that is also based Pix2PixHD \cite{pix2pixhd} network to generate anomaly image but only with edge maps and can't directly provide anomaly heatmaps. It is worth noticing that we trained four LogicAL models for four datasets evaluation. 
Comparing to LogicAL \cite{logical} (IS 2.00/\textcolor{gray}{LPIPSx10 1.58}), our generator produces (IS 4.42/\textcolor{gray}{LPIPSx10 2.58}) more realistic and diversity images and we achieve average 2.42 higher IS and 0.10 higher LPIPS. 
To compare BootGenerator and FlareGenerator, we conduct experiments using same reference color images. Comparing to BootGenerator \underline{(IS 3.96/\textcolor{gray}{LPIPSx10 2.26})}, FlareGenerator \underline{(IS 4.62/\textcolor{gray}{LPIPSx10 2.72})} achieves 0.66 higher IS and 0.05 higher LPIPS.
Fig.\ref{fig:fig6} and Fig.\ref{fig:fig7} show examples of anomaly images generated by LogicAL\cite{logical}, BootGenerator and FlareGenerator.

Table \ref{table:3} illustrates the comparison of our method and existing SOTA methods on MVTecAD \cite{mvtec} dataset. From Table \ref{table:2} we can see that the baseline method LogicAL\cite{logical} can generate competitive realistic anomalies but with very limited diversity. The proposed FlareGenerator largely increase the generation diversity and achieve the highest IS score. AnomalyDiffusion \cite{anoDiffusion2024} is a few-shot diffusion model that is particularly trained on MVTecAD \cite{mvtec} with one-third of the real anomaly data. Despite of its highest LPIPS score, it needs to train four more specialists to handle the other four datasets. On the contrary, we achieve competitive diversity and use only one generative GAN-based model that works for five datasets and is scalable for handling more categories.

\begin{table*}[t]
\begin{center}
\setlength{\abovecaptionskip}{0.cm}
\caption{Generation quality comparison of FlareGenerator with existing SOTA methods on MVTecAD \cite{mvtec}. } \label{table:3}
\resizebox{1.0\textwidth}{26mm}{
\begin{tabular}{c|ccccccc|c}

  \hline
  \multirow{2}{*}{\textbf{Category}}

  & \multicolumn{1}{c}{\textbf{DiffAug\cite{diffaug}}}  
  
  & \multicolumn{1}{c}{\textbf{CropPaste\cite{croppaste}}} 
  & \multicolumn{1}{c}{\textbf{SDGAN\cite{slsg}}  }  
  & \multicolumn{1}{c}{\textbf{DGAN\cite{defectGAN2021}}}
  & \multicolumn{1}{c}{\textbf{DFMGAN\cite{DFMGAN2023}}}  
  & \multicolumn{1}{c}{\textbf{AnoDiffusion\cite{anoDiffusion2024}}}
  & \multicolumn{1}{c|}{\textbf{LogicAL\cite{logical}}}
  & \multicolumn{1}{c}{\textbf{Ours}}
  \\
  
  \cline{2-9} 
  & \multicolumn{8}{c}{IS$\uparrow$ (Inception Score) / \textcolor{gray}{LPIPS$\uparrow$ (Learned Perceptual Image Patch Similarity)}}
  \\
    
    \hline
  bottle 
  		&1.59/\textcolor{gray}{0.03} 	  		 
  
  		&1.43/\textcolor{gray}{0.04}
  		&1.57/\textcolor{gray}{0.06}  
  		&1.39/\textcolor{gray}{0.07}
  		&1.62/\textcolor{gray}{0.12}
  		&1.58/\textcolor{gray}{0.19}
  		&1.43/\textcolor{gray}{0.11}
  		&\textbf{3.25}/\textbf{\textcolor{gray}{0.24}} 
  		\\
  		
  cable 
		&1.72/\textcolor{gray}{0.07}   		  		 
  	
  		&1.74/\textcolor{gray}{0.25}    
  		&1.89/\textcolor{gray}{0.19}  
  		&1.70/\textcolor{gray}{0.22}
  		&1.96/\textcolor{gray}{0.25}
  		&2.13/\textbf{\textcolor{gray}{0.41}}
  		&2.13/\textcolor{gray}{0.22} 
  		&\textbf{3.95}/\textcolor{gray}{0.29} 
  		\\
  		
  capsule 
		&1.34/\textcolor{gray}{0.03}

  		&1.23/\textcolor{gray}{0.05}
  		&1.49/\textcolor{gray}{0.03} 
  		&1.59/\textcolor{gray}{0.04}
  		&1.59/\textcolor{gray}{0.11}
  		&1.59/\textcolor{gray}{0.21}
  		&2.07/\textcolor{gray}{0.05}
  		&\textbf{4.94}/\textbf{\textcolor{gray}{0.25}} 
  		\\
  		
  carpet 
		&1.19/\textcolor{gray}{0.06}

  		&1.17/\textcolor{gray}{0.11}
  		&1.18/\textcolor{gray}{0.11}  
  		&1.24/\textcolor{gray}{0.12}
  		&1.23/\textcolor{gray}{0.13}
  		&1.16/\textbf{\textcolor{gray}{0.24}}
  		&1.37/\textcolor{gray}{0.14}
  		&\textbf{4.56}/\textcolor{gray}{0.22} 
  		\\
  		
  grid 
		&1.96/\textcolor{gray}{0.06}  
  		
  	   &2.00/\textcolor{gray}{0.12} 
  	   &1.95/\textcolor{gray}{0.10} 
  	   &2.01/\textcolor{gray}{0.12} 
  	   &1.97/\textcolor{gray}{0.13}
  	   &2.04/\textbf{\textcolor{gray}{0.44}}
  	   &2.60/\textcolor{gray}{0.17}
  	   &\textbf{4.81}/\textcolor{gray}{0.19} 
  	   \\
  	   
  hazelnut 
		&1.67/\textcolor{gray}{0.05}

  		&1.74/\textcolor{gray}{0.21} 
  		&1.85/\textcolor{gray}{0.16} 
  		&1.87/\textcolor{gray}{0.19}
  		&1.93/\textcolor{gray}{0.24}
  		&2.13/\textcolor{gray}{0.31}
  		&2.20/\textcolor{gray}{0.20}
  		&\textbf{4.39}/\textbf{\textcolor{gray}{0.27}} 
  		\\
  		
  leather 
		&2.07/\textcolor{gray}{0.06}

  		&1.47/\textcolor{gray}{0.14}
  		&2.04/\textcolor{gray}{0.12} 
  		&2.12/\textcolor{gray}{0.14}
  		&2.06/\textcolor{gray}{0.17}
  		&1.94/\textbf{\textcolor{gray}{0.41}}
  		&1.58/\textcolor{gray}{0.16}
  		&\textbf{4.28}/\textcolor{gray}{0.22} 
  		\\
  		
  metal nut 
		&1.58/\textcolor{gray}{0.29}

  		&1.56/\textcolor{gray}{0.15}
  		&1.45/\textcolor{gray}{0.28}  
  		&1.47/\textcolor{gray}{0.30}
  		&1.49/\textcolor{gray}{0.32}
  		&1.96/\textbf{\textcolor{gray}{0.30}}
  		&1.86/\textcolor{gray}{0.16}
  		&\textbf{4.26}/\textcolor{gray}{0.27} 
  		\\
  		
  pill 
	&1.53/\textcolor{gray}{0.05}

  	&1.49/\textcolor{gray}{0.11}  
  	&1.61/\textcolor{gray}{0.07}
  	&1.61/\textcolor{gray}{0.10} 
  	&1.63/\textcolor{gray}{0.16}
  	&1.61/\textbf{\textcolor{gray}{0.26}}
  	&2.01/\textcolor{gray}{0.14}
  	&\textbf{4.45}/\textcolor{gray}{0.21} 
  	\\
  	
  screw 
	&1.10/\textcolor{gray}{0.10}

  	&1.12/\textcolor{gray}{0.16} 
  	&1.17/\textcolor{gray}{0.10} 
  	&1.19/\textcolor{gray}{0.12}
  	&1.12/\textcolor{gray}{0.14}
  	&1.28/\textcolor{gray}{0.30}
  	&1.58/\textcolor{gray}{0.05}
  	&\textbf{4.88}/\textbf{\textcolor{gray}{0.32}} 
  	\\
  	 
  tile 
  	&1.93/\textcolor{gray}{0.09}

  	&1.83/\textcolor{gray}{0.20} 
  	&2.53/\textcolor{gray}{0.21} 
  	&2.35/\textcolor{gray}{0.22}
  	&2.39/\textcolor{gray}{0.22}
  	&2.54/\textbf{\textcolor{gray}{0.55}}
  	&1.72/\textcolor{gray}{0.04}
	&\textbf{4.60}/\textcolor{gray}{0.20} 
  	\\
  	
  toothbrush 
	&1.33/\textcolor{gray}{0.06}

  	&1.30/\textcolor{gray}{0.08}
  	&1.78/\textcolor{gray}{0.03}  
  	&1.85/\textcolor{gray}{0.03}
  	&1.82/\textcolor{gray}{0.18}
  	&1.68/\textcolor{gray}{0.21}
  	&1.45/\textcolor{gray}{0.24}
  	&\textbf{4.03}/\textbf{\textcolor{gray}{0.25}} 
  	\\
  	
  transistor 
	&1.34/\textcolor{gray}{0.05}

  	&1.39/\textcolor{gray}{0.15}
  	&1.76/\textcolor{gray}{0.13}  
  	&1.47/\textcolor{gray}{0.13}
  	&1.64/\textcolor{gray}{0.25}
  	&1.57/\textbf{\textcolor{gray}{0.34}}
  	&1.47/\textcolor{gray}{0.18}
  	&\textbf{3.22}/\textcolor{gray}{0.28} 
  	\\
  	
  wood 
	&2.05/\textcolor{gray}{0.30}

  	&1.95/\textcolor{gray}{0.23}
  	&2.12/\textcolor{gray}{0.25}  
  	&2.19/\textcolor{gray}{0.29}
  	&2.12/\textcolor{gray}{0.35}
  	&2.33/\textbf{\textcolor{gray}{0.37}}
  	&1.76/\textcolor{gray}{0.07}
  	&\textbf{3.90}/\textcolor{gray}{0.21}
  	\\
  	
  zipper 
	&1.30/\textcolor{gray}{0.05}

  	&1.23/\textcolor{gray}{0.11} 
  	&1.25/\textcolor{gray}{0.10}  
  	&1.25/\textcolor{gray}{0.10}
  	&1.29/\textbf{\textcolor{gray}{0.27}}
  	&1.39/\textcolor{gray}{0.25}
  	&1.55/\textcolor{gray}{0.21}
  	&\textbf{4.12}/\textbf{\textcolor{gray}{0.27}} 
  	\\
  	
  \hline
  Mean 
	&1.58/\textcolor{gray}{0.09}

  	&1.51/\textcolor{gray}{0.14} 
  	&1.71/\textcolor{gray}{0.13}  
  	&1.69/\textcolor{gray}{0.15}
  	&1.72/\textcolor{gray}{0.20}
  	&1.80/\textbf{\textcolor{gray}{0.32}}
  	&1.78/\textcolor{gray}{0.14}
  	&\textbf{4.24}/\textcolor{gray}{0.25} 
  	\\
  \hline
\end{tabular}
	}
	
\vspace{-0.5cm}
\end{center}
\end{table*}

\subsection{Ablation Study}

\begin{figure}[tb]
  \centering
  \begin{minipage}{0.4\textwidth}
  \centering
      \includegraphics[height=3.cm]{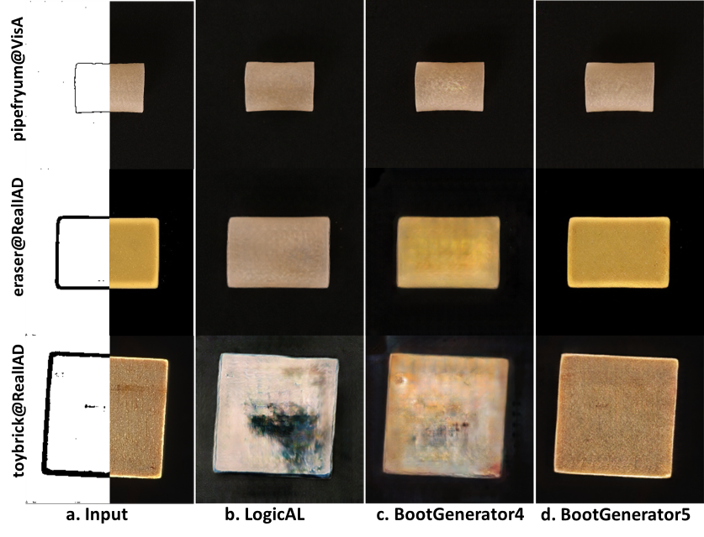}
    \caption{Comparison of scalability.}
    \label{fig:fig5}
  \end{minipage}
\hfill
  \begin{minipage}{0.58\textwidth}
    \includegraphics[height=3.cm]{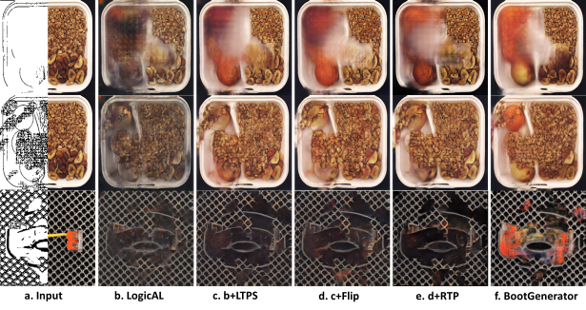}
    \caption{Comparison of augmentations.}
    \label{fig:fig6}
   \end{minipage}

\end{figure}

\begin{figure}[tb]
\centering
    \includegraphics[height=3.5cm]{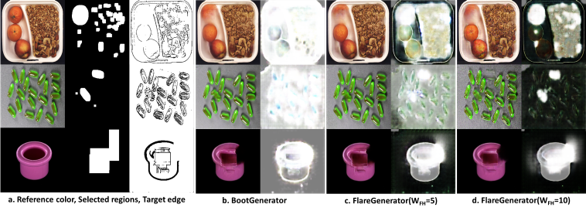}
    \caption{Comparison of generated anomaly(left) and heatmap(right).}
    \label{fig:fig7}
\end{figure}

\begin{figure}[tb]
\centering
    \includegraphics[height=3.cm]{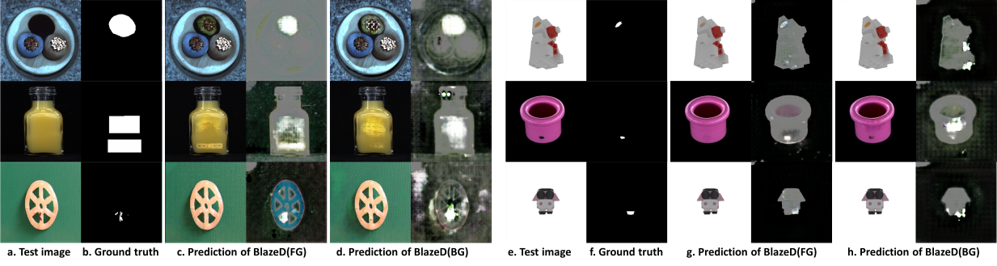}
    \caption{Comparison BlazeDetectors supervised by FlareGenerator(FG) and BootGenerator(BG).}
    \label{fig:fig8}
\end{figure}

\textbf{Scalability.}
Fig.\ref{fig:fig5} illustrates the scalability of our BootGenerator comparing with baseline method LogicAL \cite{logical}. As shown in Fig.\ref{fig:fig5}a, three test objects from different datasets share similar edge maps. From Fig.\ref{fig:fig5}b to Fig.\ref{fig:fig5}d, LogicAL \cite{logical}, BootGenerator4 and BootGenerator5 models are trained on \{VisA\} , \{VisA, MVTecAD, MVTecLOCO, MADSim\} and \{VisA, MVTecAD, MV-TecLOCO, MADSim, RealIAD\} respectively. Due to only use an edge map as input, LogicAL \cite{logical} struggles in generating remembered content(pipefryum) for unseen edge maps(eraser). On the contrary, BootGenerator4 generates images that are more close to the input reference color image. By adding samples from RealIAD datasets for training, BootGenerator5 scales its generation ability and outputs more authentic results.

\noindent\textbf{Augmentations.} 
Fig.\ref{fig:fig6} demonstrates the effectiveness of three augmentations we used for training BootGenerator, including LTPS(Fig.\ref{fig:fig6}c: Local TPS), Flip(Fi-g.\ref{fig:fig6}d: Top-bottom/Left-right flip) and RTP(Fig.\ref{fig:fig6}e: Resize-Translation-Padding). Fig.\ref{fig:fig6}a shows examples that the edge maps are drastically edited to forge anomaly edges. As shown in Fig.\ref{fig:fig6}b, these edge maps cause mode collapse to the LogicAL \cite{logical}. By using augmentations of TPS, Flip and RTP during training, the LogicAL \cite{logical} can generate much better results. By comparison, BootGenerator(Fig.\ref{fig:fig6}f) obtains high-fidelity results especially in the unedited pixels, such as the orange region.

\noindent\textbf{Anomaly heatmaps.}
Fig.\ref{fig:fig7} shows the necessity of FlareGenerator for generating more accurate anomaly heatmaps. As shown in Fig.\ref{fig:fig7}a, the anomalies should happen in the selected regions(Semantic or Stochastic) according to the target edge maps. Even though BootGenerator produces reasonable anomaly images, the corresponding anomaly heatmaps(Fig.\ref{fig:fig7}b-right) barely provide useful information. By learning from BootGenerator and region masks(Fig.\ref{fig:fig7}a-middle), FlareGenerator outputs more accurate anomaly heatmaps(Fig.\ref{fig:fig7}d-right) that align to the anomalies. Moreover, Fig.\ref{fig:fig7}c shows that the weight of anomaly heatmap loss is preferred to be set as 10.

\noindent\textbf{Anomaly detection.}
Fig.\ref{fig:fig8} illustrates the effectiveness of using FlareGenerator rather than BootGenerator to supervise the training of BlazeDetector. 
Comparing to Fig.\ref{fig:fig8}d and g, the BlazeGenerator(Fig.\ref{fig:fig8}c,g) supervised by FlareGenerator produces higher quality normal images(left) and anomaly heatmaps(right).

\section{Conclusion}
This paper proposes a novel scalable framework that unifies unsupervised anomaly generation and anomaly localization. With same network architecture, we trained two consecutive anomaly generators(BootGenerator and FlareGenerator) and one anomaly predictor(BlazeDetector) for 82 categories of 5 datasets. The BootGenerator triggers the generation ability of absorbing structure of a target edge map and appearance of a reference color image. Meanwhile, it learns a heatmap that is further improved by subsequent FlareGenerator to more accurately indicate pixel-level anomaly locations of the generated anomaly images. By simply swapping normal images with the generated anomaly images for training, we learn an anomaly predictor(BlazeDetector) that directly outputs anomaly heatmaps. Even thought the performance of BlazeDetector has room to improve, this work suggests a high potential for simplifying existing pipelines of unsupervised learning of anomaly generation and localization. Improving its zero-shot generation ability is a promising direction for future work.



%
%
\bibliographystyle{splncs04}
\bibliography{main}

\begin{figure}[htbp]
  \centering
  \includegraphics[height=9cm]{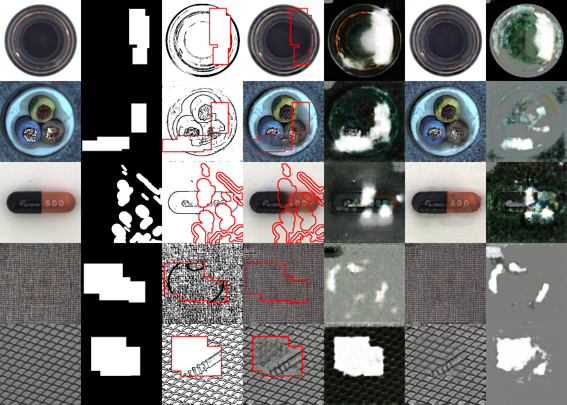}
  \caption{From left to right: Normal image, (Selected regions, Anomaly edges, Anomaly images, Anomaly heatmap) generated by FlareGenerator, (Normal image, Anomaly heatmap) generated by BlazeDetector on test/good of MVTecAD \cite{mvtec} part1/3.}
  \label{fig:mvtec2d0}
\end{figure}

\begin{figure}[htbp]
  \centering
  \includegraphics[height=9cm]{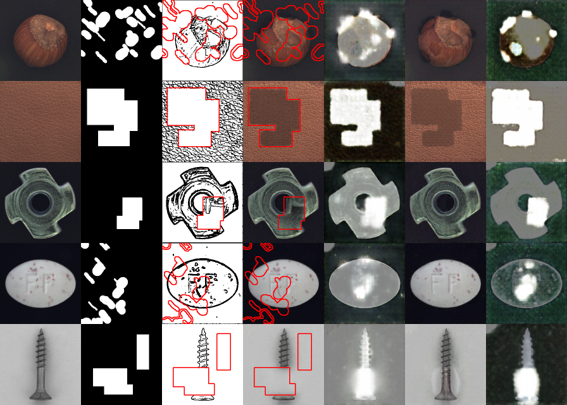}
  \caption{From left to right: Normal image, (Selected regions, Anomaly edges, Anomaly images, Anomaly heatmap) generated by FlareGenerator, (Normal image, Anomaly heatmap) generated by BlazeDetector on test/good of MVTecAD \cite{mvtec} part2/3.} 

  \label{fig:mvtec2d1}
\end{figure}

\begin{figure}[htbp]
  \centering
  \includegraphics[height=9cm]{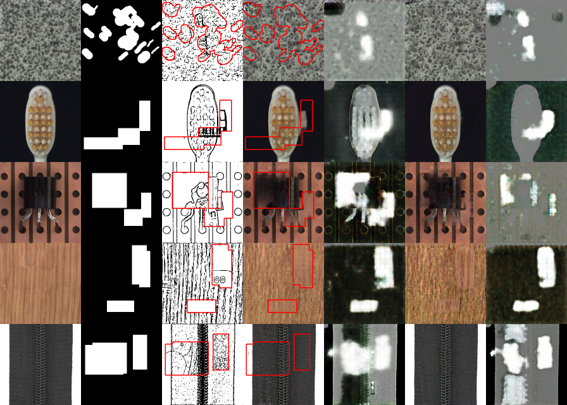}
  \caption{From left to right: Normal image, (Selected regions, Anomaly edges, Anomaly images, Anomaly heatmap) by FlareGenerator, (Normal image, Anomaly heatmap) by BlazeDetector on test/good of MVTecAD \cite{mvtec} part3/3. 
}
  \label{fig:mvtec2d2}
\end{figure}

\begin{figure}[htbp]
  \centering
  \includegraphics[height=7cm]{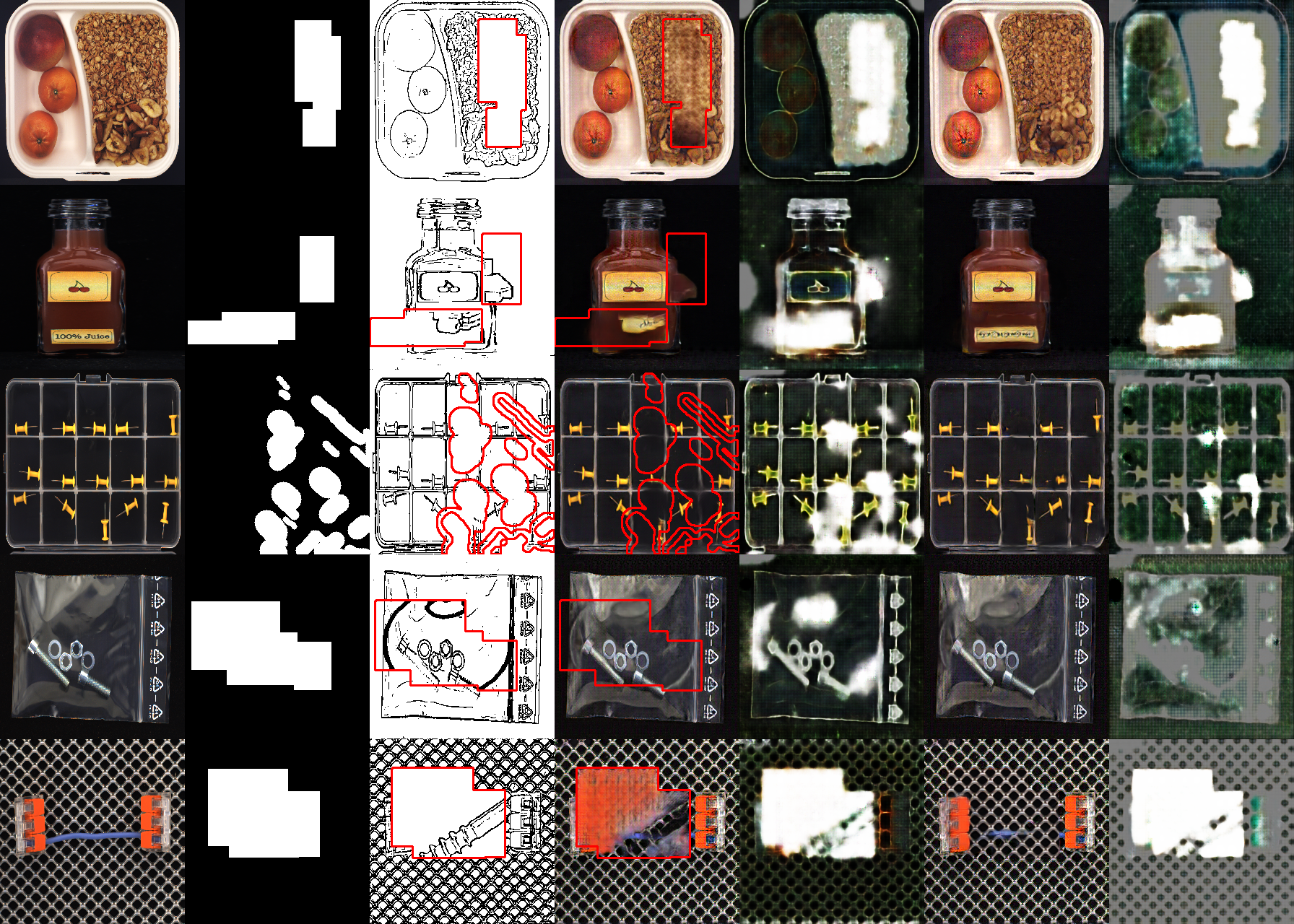}
  \caption{From left to right: Normal image, (Selected regions, Anomaly edges, Anomaly images, Anomaly heatmap) generated by FlareGenerator, (Normal image, Anomaly heatmap) generated by BlazeDetector on test/good of MVTecLOCO \cite{mvtecloco}. 
}
  \label{fig:mvtecloco}
\end{figure}

\begin{figure}[htbp]
  \centering
  \includegraphics[height=7cm]{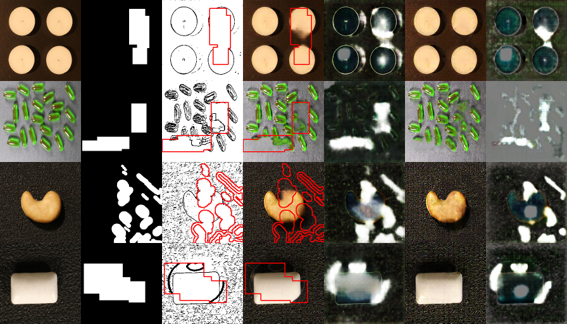}
  \caption{From left to right: Normal image, (Selected regions, Anomaly edges, Anomaly images, Anomaly heatmap) generated by FlareGenerator, (Normal image, Anomaly heatmap) generated by BlazeDetector on test/good of VisA \cite{visa} part1/3.
}
  \label{fig:visa0}
\end{figure}

\begin{figure}[htbp]
  \centering
  \includegraphics[height=7cm]{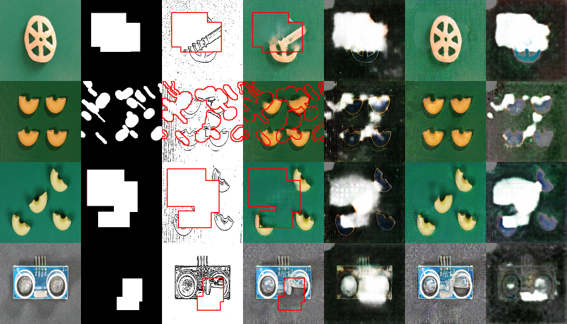}
  \caption{From left to right: Normal image, (Selected regions, Anomaly edges, Anomaly images, Anomaly heatmap) generated by FlareGenerator, (Normal image, Anomaly heatmap) generated by BlazeDetector on test/good of VisA \cite{visa} part2/3. 
}
  \label{fig:visa1}
\end{figure}

\begin{figure}[htbp]
  \centering
  \includegraphics[height=7cm]{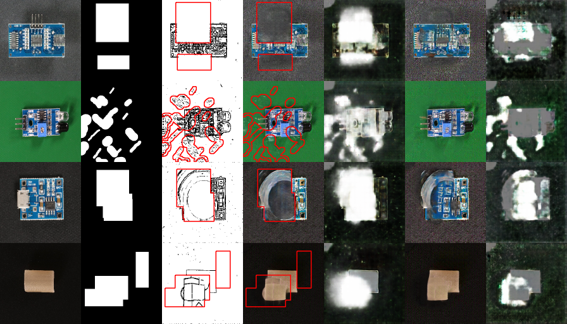}
  \caption{From left to right: Normal image, (Selected regions, Anomaly edges, Anomaly images, Anomaly heatmap) generated by FlareGenerator, (Normal image, Anomaly heatmap) generated by BlazeDetector on test/good of VisA \cite{visa} part3/3.
}
  \label{fig:visa2}
\end{figure}

\begin{figure}[htbp]
  \centering
  \includegraphics[height=7cm]{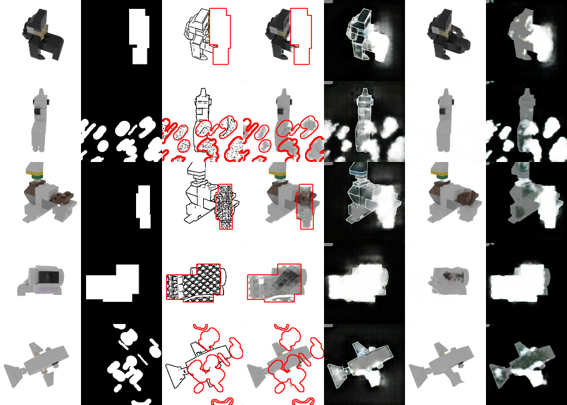}
  \caption{From left to right: Normal image, (Selected regions, Anomaly edges, Anomaly images, Anomaly heatmap) generated by FlareGenerator, (Normal image, Anomaly heatmap) generated by BlazeDetector on test/good of MADSim \cite{pad} part1/4.
}
  \label{fig:madsim0}
\end{figure}

\begin{figure}[htbp]
  \centering
  \includegraphics[height=7cm]{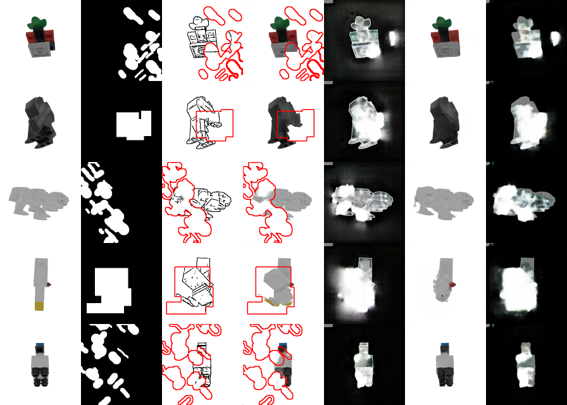}
  \caption{From left to right: Normal image, (Selected regions, Anomaly edges, Anomaly images, Anomaly heatmap) generated by FlareGenerator, (Normal image, Anomaly heatmap) generated by BlazeDetector on test/good of MADSim \cite{pad} part2/4.
}
  \label{fig:madsim1}
\end{figure}

\begin{figure}[htbp]
  \centering
  \includegraphics[height=7cm]{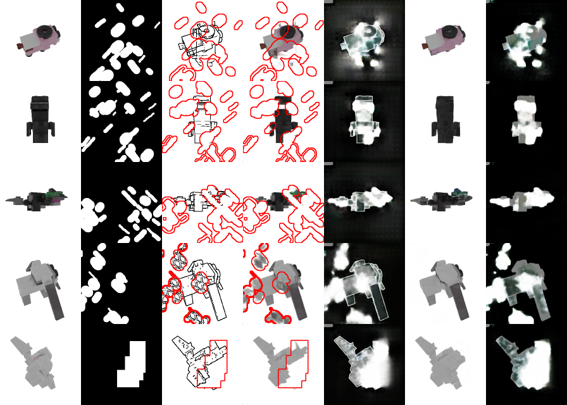}
  \caption{From left to right: Normal image, (Selected regions, Anomaly edges, Anomaly images, Anomaly heatmap) generated by FlareGenerator, (Normal image, Anomaly heatmap) generated by BlazeDetector on test/good of MADSim \cite{pad} part3/4.
}
  \label{fig:madsim2}
\end{figure}

\begin{figure}[htbp]
  \centering
  \includegraphics[height=7cm]{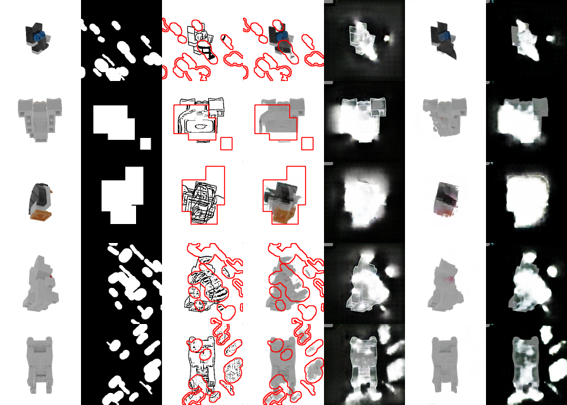}
  \caption{From left to right: Normal image, (Selected regions, Anomaly edges, Anomaly images, Anomaly heatmap) generated by FlareGenerator, (Normal image, Anomaly heatmap) generated by BlazeDetector on test/good of MADSim \cite{pad} part4/4.
}
  \label{fig:madsim3}
\end{figure}

\begin{figure}[htbp]
  \centering
  \includegraphics[height=7cm]{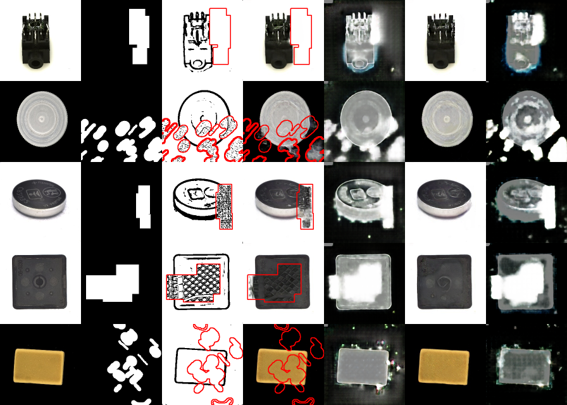}
  \caption{From left to right: Normal image, (Selected regions, Anomaly edges, Anomaly images, Anomaly heatmap) generated by FlareGenerator, (Normal image, Anomaly heatmap) generated by BlazeDetector on test/good of RealIAD \cite{realiad} part1/6.
}
  \label{fig:realiad0}
\end{figure}

\begin{figure}[htbp]
  \centering
  \includegraphics[height=7cm]{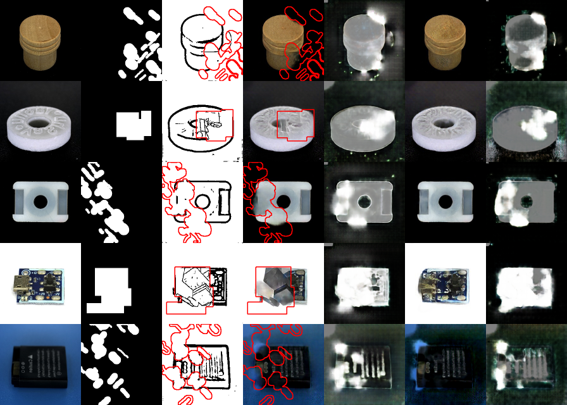}
  \caption{From left to right: Normal image, (Selected regions, Anomaly edges, Anomaly images, Anomaly heatmap) generated by FlareGenerator, (Normal image, Anomaly heatmap) generated by BlazeDetector on test/good of RealIAD \cite{realiad} part2/6.
}
  \label{fig:realiad1}
\end{figure}

\begin{figure}[htbp]
  \centering
  \includegraphics[height=7cm]{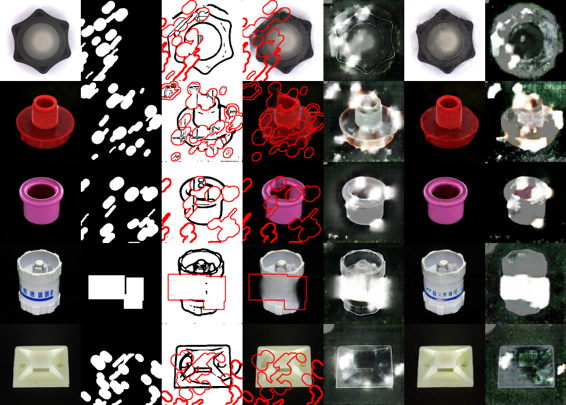}
  \caption{From left to right: Normal image, (Selected regions, Anomaly edges, Anomaly images, Anomaly heatmap) generated by FlareGenerator, (Normal image, Anomaly heatmap) generated by BlazeDetector on test/good of RealIAD \cite{realiad} part3/6.
}
  \label{fig:realiad2}
\end{figure}

\begin{figure}[htbp]
  \centering
  \includegraphics[height=7cm]{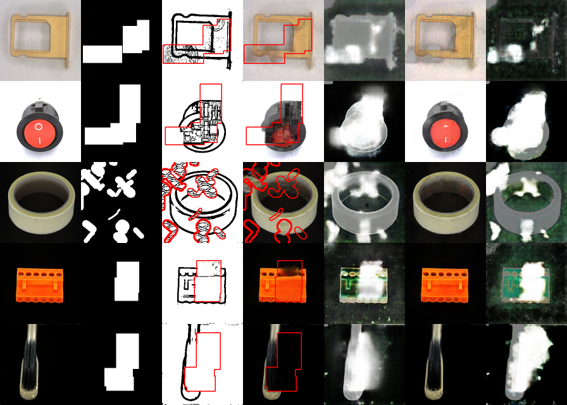}
  \caption{From left to right: Normal image, (Selected regions, Anomaly edges, Anomaly images, Anomaly heatmap) generated by FlareGenerator, (Normal image, Anomaly heatmap) generated by BlazeDetector on test/good of RealIAD \cite{realiad} part4/6.
}
  \label{fig:realiad3}
\end{figure}

\begin{figure}[htbp]
  \centering
  \includegraphics[height=7cm]{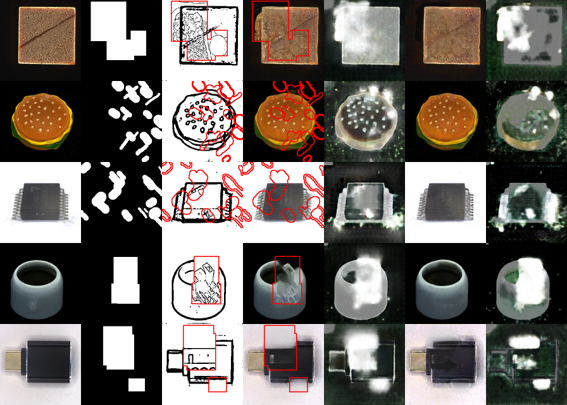}
  \caption{From left to right: Normal image, (Selected regions, Anomaly edges, Anomaly images, Anomaly heatmap) generated by FlareGenerator, (Normal image, Anomaly heatmap) generated by BlazeDetector on test/good of RealIAD \cite{realiad} part5/6.
}
  \label{fig:realiad4}
\end{figure}

\begin{figure}[htbp]
  \centering
  \includegraphics[height=7cm]{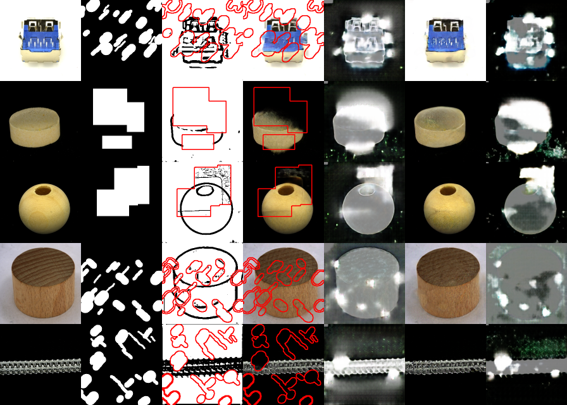}
  \caption{From left to right: Normal image, (Selected regions, Anomaly edges, Anomaly images, Anomaly heatmap) generated by FlareGenerator, (Normal image, Anomaly heatmap) generated by BlazeDetector on test/good of RealIAD \cite{realiad} part6/6. 
}
  \label{fig:realiad5}
\end{figure}
\end{document}